\documentclass[lettersize,journal]{IEEEtran}
\usepackage{amsmath,amsfonts}
\usepackage{algorithmic}
\usepackage{algorithm}
\usepackage{array}
\usepackage{textcomp}
\usepackage{stfloats}
\usepackage{url}
\usepackage{verbatim}
\usepackage{graphicx}
\usepackage{cite}
\usepackage{subcaption}
\usepackage{multirow}
\def\ourmodel{PLML}

\newcommand{\tr}[1]{\textcolor{black}{#1}}
\newcommand{\rec}[1]{\textcolor{black}{#1}}
\newcommand{\red}{\textcolor{red}}
\usepackage[pagebackref=true,breaklinks=true,letterpaper=true,colorlinks,bookmarks=false]{hyperref}

\usepackage{cleveref}
\begin{document}

\title{Pseudo-Labeling Based Practical Semi-Supervised Meta-Training for Few-Shot Learning}

\author{Xingping Dong, Tianran Ouyang, Shengcai Liao, Bo Du, and Ling Shao,~\IEEEmembership{Fellow,~IEEE}
\IEEEcompsocitemizethanks{
%
\IEEEcompsocthanksitem X. Dong, T. Ouyang and B. Du are with the School of Computer Science, National Engineering Research Center for Multimedia Software, Institute of Artificial Intelligence, Hubei Key Laboratory of Multimedia and Network Communication Engineering, Wuhan University, Wuhan 430072, China. 
(Email: xingpingdong@whu.edu.cn, ouyangtianran@whu.edu.cn and dubo@whu.edu.cn)
\IEEEcompsocthanksitem S. Liao is with the College of Information Technology, United Arab Emirates University, Al Ain, UAE.
(Email: scliao@ieee.org)
%
\IEEEcompsocthanksitem L. Shao is with the UCAS-Terminus AI Lab, University of Chinese Academy of Sciences, Beijing, 100049, China. (email: ling.shao@ieee.org) 
\IEEEcompsocthanksitem This work is partially funded by the Fundamental Research Funds for the Central Universities (Ref. No.: 2042022kf1198), the Science and Technology Development Fund (001/2024/SKL) and the State Key Laboratory of Internet of Things for Smart City (University of Macau) (Ref. No.: SKL-IoTSC(UM)-2024-2026/ORP/GA04/2023), and National Natural Science Foundation of China (Grant No. 62471342)
\IEEEcompsocthanksitem Corresponding author: \textit{Xingping Dong} and \textit{Shengcai Liao}
%
}
}




\maketitle

\begin{abstract}
Most existing few-shot learning (FSL) methods require a large amount of labeled data in meta-training, which is a major limit. To reduce the requirement of labels, a semi-supervised meta-training (SSMT) setting has been proposed for FSL, which includes only a few labeled samples and numbers of unlabeled samples in base classes. However, existing methods under this setting require class-aware sample selection from the unlabeled set, which violates the assumption of unlabeled set. In this paper, we propose a practical semi-supervised meta-training setting with truly unlabeled data to facilitate the applications of FSL in realistic scenarios. To better utilize both the labeled and truly unlabeled data, we propose a simple and effective meta-training framework, called pseudo-labeling based meta-learning (PLML). Firstly, we train a classifier via common semi-supervised learning (SSL) and use it to obtain the pseudo-labels of unlabeled data. Then we build few-shot tasks from labeled and pseudo-labeled data and design a novel finetuning method with feature smoothing and noise suppression to better learn the FSL model from noise labels. Surprisingly, through extensive experiments across two FSL datasets, we find that this simple meta-training framework effectively prevents the performance degradation of various FSL models under limited labeled data, and also significantly outperforms the representative SSMT models. Besides, benefiting from meta-training, {our method also improves several representative SSL algorithms as well. We provide the training code and usage examples at https://github.com/ouyangtianran/PLML.}
\end{abstract}


\section{Introduction}

\label{sec:intro}

\begin{figure}[!htb]
\centering
  \begin{subfigure}{0.49\linewidth}
\includegraphics[width = 1. \textwidth]{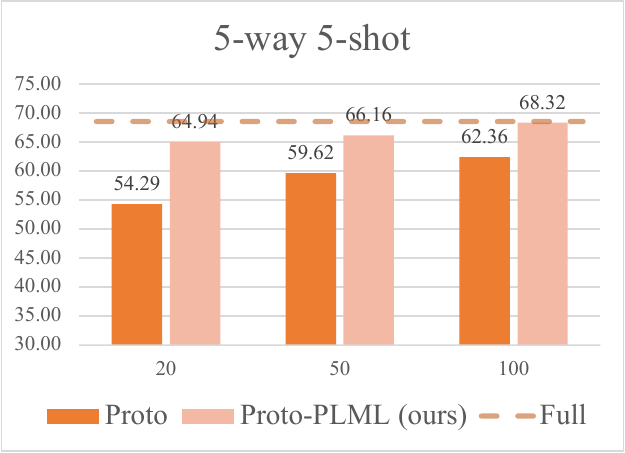}
    \caption{Inductive}
    \label{fig:1-in}
  \end{subfigure}
  \hfill
\begin{subfigure}{0.49\linewidth}

\includegraphics[width = 1. \textwidth]{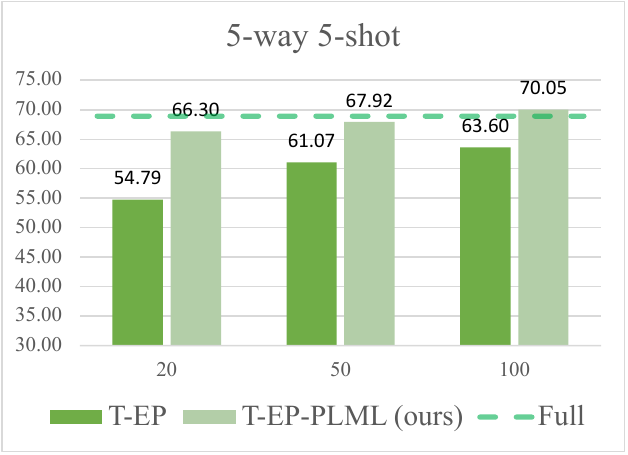}
    \caption{Transductive}
    \label{fig:1-trans}
  \end{subfigure}
  \hfill
\begin{subfigure}{0.49\linewidth}

\includegraphics[width = 1. \textwidth]{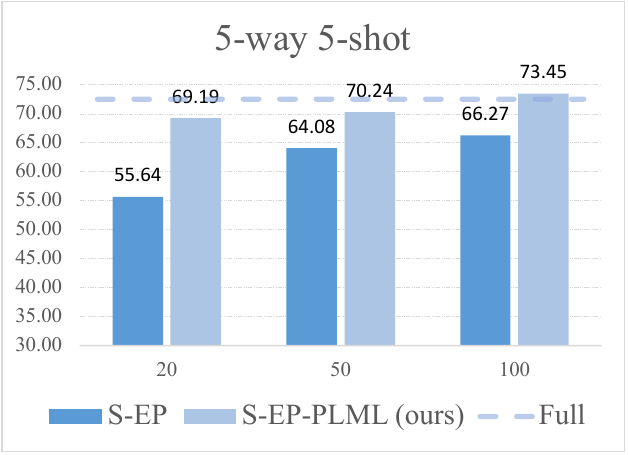}
    \caption{Semi-Supervised}
    \label{fig:1-semi}
  \end{subfigure}
  \hfill  
\begin{subfigure}{0.49\linewidth}

\includegraphics[width = 1. \textwidth]{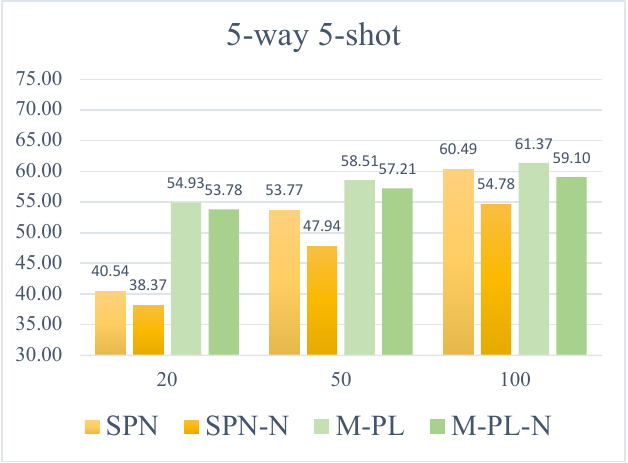}
    \caption{Meta-Training Based}
    \label{fig:1-meta}
  \end{subfigure}

\caption{
\textbf{Illustrations of performance degradation with the reduction of labels in few-shot learning (FSL).} The horizontal and vertical axes represent the number of labels per class and accuracy on \textit{mini}ImageNet~\cite{ravi2016optimization}. 
{We demonstrate the performance degradation of three representative Few-Shot Learning (FSL) models (inductive \textit{Proto} (a), transductive \textit{T-EP} \cite{rodriguez2020embedding} (b), and semi-supervised \textit{S-EP} \cite{rodriguez2020embedding} (c)) when using different numbers of training labels in various tasks. "Full" denotes training with fully labeled data, while "PLML" indicates incorporation with our pseudo-labeling based meta-learning approach. In (d), we compare two models: \textit{SPN} \cite{ren2018meta}, and \textit{M-PL} \cite{li2022platinum}, in both the original setting and our new setting (*-N).}
}\label{fig:p-degrade}
\end{figure}

\begin{figure*}[!htb]
\centering
\includegraphics[width = 1\textwidth]{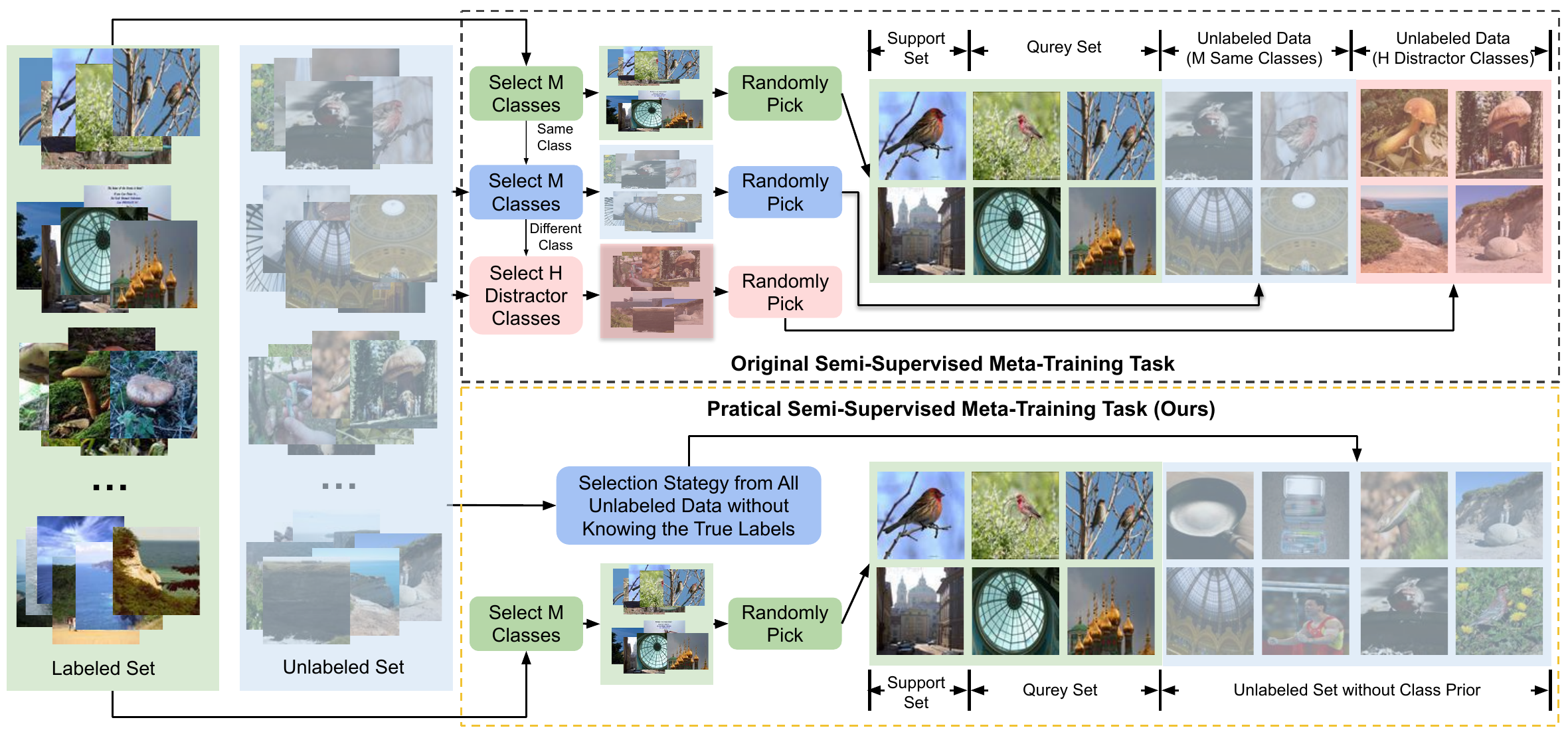} \\

\caption{
\textbf{Existing semi-supervised meta-training task \cite{ren2018meta} (black box) \textit{v.s.} the proposed (yellow box).} Green and blue blocks represent labeled and unlabeled data, respectively. 
}\label{fig:ss-meta-train}
\end{figure*}

\IEEEPARstart{R}{ecently}, {researchers pay much attention to few-shot learning (FSL), which aims to recognize novel (unseen) objects by only learning from one or a few examples (support set) in each category, with some exploration and accumulation of related techniques~\cite{song2023comprehensive,li2023libfewshot}.}
Existing FSL methods are grouped into three types in terms of the inference manner: (i) \tr{the traditional inductive FSL methods~\cite{finn2017model,snell2017prototypical} (IFSL) do not use additional information during inference and evaluate the test data one by one;} (ii) the transductive inference \cite{joachims1999transductive} for FSL (TFSL)~\cite{liu2018learning,qiao2019transductive,rodriguez2020embedding} supposes that we can utilize all test data to learn the model for inference; (iii) Semi-supervised FSL (SSFSL)~\cite{ren2018meta,li2019learning,rodriguez2020embedding,lazarou2021iterative,li2022platinum,ye2021task} assumes we can have additional unlabeled data for inference. 
\tr{\noindent\textbf{Limitation 1: heavy reliance on base dataset with many labeled data.}
Although they achieve impressive performance on novel classes with few labels,} most FSL methods still need to pre-train their models on a disjoint but relevant base dataset with many labeled samples, even for SSFSL approaches.
Specifically, IFSL and TFSL methods assume that they have access to {abundant labeled examples} in the base dataset for training. The SSFSL methods based on transfer-learning \cite{rodriguez2020embedding,ling2022semi} also use all base labels for pre-training, 
\tr{and employ the semi-supervised inference on the novel dataset.}
\tr{To investigate the impact of limitation 1, we reduce the number of labeled data and retrain three representative FSL models with different inference manners} (inductive Proto~\cite{snell2017prototypical}, transductive T-EP, and semi-supervised S-EP~\cite{rodriguez2020embedding}), which require full labels during meta-training. 
As shown in Fig.~\ref{fig:p-degrade}, they suffer from obvious \textit{performance degradation with label reduction}.
This indicates that most FSL methods still heavily rely on a mass of labeled data, which is one major limit for the applications with expensive data annotation (\emph{e.g.} medical data).

\tr{Unlike transfer-learning based SSFSL methods\cite{rodriguez2020embedding,ling2022semi}, which use semi-supervised inference only on the novel dataset,} the meta-training based SSFSL methods split the base dataset into labeled and unlabeled sets, and design new meta-learning methods to reduce the requirement of labeled data and fully utilize the unlabeled data. 
As shown in Fig.~\ref{fig:1-meta}, we evaluate two representative models: SPN~\cite{ren2018meta}, and M-PL~\cite{li2022platinum}, with different numbers of labeled data.
\tr{We observe that the evaluation scores still drop sharply with the reduction of labeled data, especially when the number of labels per class is extremely small (\emph{e.g.} 20).} Furthermore, meta-training based SSFSL models have \textit{large gaps of accuracy} compared with transfer-learning based SSFSL methods under the same limited conditions(\emph{e.g.} M-PL in Fig.~\ref{fig:1-meta} \emph{vs.} S-EP in Fig.~\ref{fig:1-semi}). 

\tr{\noindent\textbf{Limitation 2: class-prior selecting strategy in meta-training based SSFSL models.}}
Existing meta-training based SSFSL models \cite{ren2018meta,li2022platinum} require a \textit{class-prior selecting strategy} for training, which is not suitable for truly unlabeled data. 
\tr{As shown in Fig.~\ref{fig:ss-meta-train}, during task construction, meta-training based SSFSL models first select M classes from both labeled and unlabeled sets, then randomly pick up several samples from these classes to construct the tasks. To ensure the selected unlabeled samples are from these M classes, they actually need to know all the labels of samples in the unlabeled set, which violates the principle of an unlabeled set in practice.}
Is this strategy necessary for SSFSL? Can we remove it? These are still unexplored questions.

\tr{\noindent\textbf{The new setting.}
To overcome the limitation 2, we propose a new and practical semi-supervised meta-training setting for FSL.}
{As shown in Fig.~\ref{fig:ss-meta-train}, our practical task only requires selecting M classes from the labeled set and using any class-not-known selection strategy to pick up unlabeled samples.}
Specifically, the base dataset is split into labeled and unlabeled sets, and we forbid the \textit{class-prior selecting strategy} for the unlabeled set to make sure we do not have access to the true labels of the unlabeled data. For the novel dataset, we adopt the same inference manners of previous works to focus on the impact of the number of labeled data. 

Accordingly, we retrain SPN and M-PL with different numbers of labels on our new setting, noted as SPN-N and M-PL-N. As shown in Fig.~\ref{fig:1-meta}, SPN-N suffers from obvious performance degradation while M-PL-N obtains a slight reduction in accuracy. \tr{This indicates that different models have different sensitivity to the \textit{class-prior selecting strategy}. In other words, we cannot simply infer their performance in a more practical new setting based on their performance in the original setting.} Thus, we use the new setting for experimental evaluation to design a more practical FSL approach. 

\noindent\textbf{The two-stage meta-training framework.}
To overcome aforementioned limitations, we propose a two-stage meta-training framework, called pseudo-labeling based meta-learning (\ourmodel), to fully use unlabeled data. In the first stage, we pre-train a base classifier on both the labeled and unlabeled sets, by using a semi-supervised learning approach. In the second stage, the pre-trained classifier firstly generates pseudo-labels of unlabeled data. Then we construct the episodic tasks from both the labeled data and pseudo-labeled data, and propose a novel finetuning method with feature smoothing and noise suppression to train the FSL model over these constructed tasks containing noisy labels. We also finetune the base classifier during the second learning stage, to explore the impact of meta-learning for semi-supervised classification. 

\tr{Our core idea is to integrate SSL with meta-learning based SSFSL to mitigate the performance degradation rising from the reduction of labeled data in the base dataset, while also avoiding the impractical \textit{class-prior selecting strategy} in task construction.}
We instantiate our {\ourmodel} with two representative meta-learning algorithms: Proto~\cite{hsu2018unsupervised} (IFSL) and EP~\cite{rodriguez2020embedding} (TFSL and SSFSL), {and compare to recent semi-supervised learning (SSL) methods, including SemCo~\cite{nassar2021all}, FlexMatch~\cite{zhang2021flexmatch} and MarginMatch~\cite{sosea2023marginmatch}}. Extensive experiments are conducted on two standard benchmarks: \textit{mini}ImageNet~\cite{ravi2016optimization}, and \textit{tiered}ImageNet~\cite{ren2018meta}.
The results demonstrate that our simple meta-training framework can significantly alleviate the \textit{performance degradation with label reduction} for the existing FSL models and narrow the \textit{huge gaps} between meta-training and transfer-learning based SSFSL methods. 
As shown in Fig.~\ref{fig:1-semi} and \ref{fig:1-meta}, our S-EP-{\ourmodel} (based on Semco) trained with only 20 labeled samples per class achieves the accuracy of 69.19\% on \textit{mini}ImageNet for 5-way 5-shot tasks, which respectively obtains huge gains of 13.55\% and 15.41\% than the original model (S-EP) and state-of-the-art (SOTA) meta-training based model (M-PL-N).
Notably, we find that our meta-learning stage can further improve the performance of recent SSL models, which indicates that combining meta-learning might be a new direction for SSL.

In summary, our key contributions include:
\begin{itemize}
	\item We propose a more practical semi-supervised meta-training setting for FSL, as a platform to facilitate the development of training with a few labeled base data for FSL models. This is the first attempt to investigate and remove the class-prior selection in SSFSL.
	\item A simple and effective semi-supervised training algorithm, named pseudo-labeling based meta-learning, is proposed to reduce the requirement of the labeled base data for most FSL models. Our approach is the first work to introduce the semi-supervised pre-training manner to meta-training based SSFSL.
	\item We find that our meta-learning strategy can further improve the performance of recent SSL models, which indicates that using meta-training could be a promising direction to develop new SSL approaches.
	\item 
	Our approach is model-agnostic and successfully incorporates SSL to FSL. We significantly improve the performance of FSL models (EP and Proto) with limited labeled base training data.
\end{itemize}

\section{Related Work}
\subsection{Few-Shot Learning (FSL)}
\noindent\textbf{Inductive.} 
Usually, inductive methods are based on meta-learning, which can be grouped as model-, optimization-, and metric-based algorithms. 
The first group attempts to extract meta knowledge by memory models, such as recurrent and memory-augmented networks \cite{santoro2016meta,ravi2016optimization}. The second one aims to learn a generative model initiation that enables fast adaption for novel tasks ~\cite{finn2017model,finn2018probabilistic,rajeswaran2019meta,bertinetto2018meta,zintgraf2019fast,lee2019meta}. The last group attempts to learn powerful representations to discriminate novel classes \cite{koch2015siamese,vinyals2016matching,snell2017prototypical,rodriguez2020embedding}. Recently, transfer-learning based methods \cite{gidaris2018dynamic,chen2019closer,tian2020rethinking} utilize the entire training set to learn a powerful representation instead of episodic training.

\noindent\textbf{Transductive. }
Common approaches combine label propagation \cite{liu2018learning} or embedding propagation \cite{rodriguez2020embedding} by smoothing embedding. Some researchers focus on using unlabeled data better, \textit{e.g.} subspace \cite{lichtenstein2020tafssl,simon2020adaptive}, optimal-transport \cite{hu2020leveraging}, and cross-attention~\cite{hou2019cross}. {Recently, accurate prototype estimation has gained attention. For example,\cite{zhang2023prototype} introduced a prototype completion based meta-learning framework, while \cite{zhu2023transductive} proposed a prototype-based label propagation method.}

\noindent\textbf{Semi-Supervised.}
Existing methods are roughly split into two groups. 1) \textit{Meta-training based}:
The pioneering work \cite{ren2018meta} proposes an adaptive prototypical network to utilize unlabeled data. The following works introduce label propagation \cite{liu2018learning} and pseudo-labeling \cite{li2019learning}. {TACO~\cite{ye2021task} adds a task-level smoothness constraint on unlabeled data, thereby smoothing the meta-model space.} Recently, the model-agnostic meta-learning \cite{finn2017model} is extended to SSFSL by selecting reliable training subsets from unlabeled data \cite{li2022platinum}. 2) \textit{Transfer-learning based}: These methods \cite{yu2020transmatch,wang2020instance,rodriguez2020embedding,lazarou2021iterative,ling2022semi} have attracted main attention due to the high accuracy. However, they require abundant base labels, limiting them for realistic scenarios.
As aforementioned, most FSL models rely heavily on many training labels. This paper aims to design a new framework to facilitate FSL training with fewer labels.

\subsection{Semi-Supervised Learning (SSL)}
Since the pioneering work, II-Model~\cite{rasmus2015semi}, pseudo-labeling and consistency regularization techniques have attracted much attention in the SSL community and generated various algorithms \cite{samuli2017temporal,tarvainen2017mean,miyato2018virtual,kuo2020featmatch}. Recently proposed models~\cite{xie2019unsupervised,berthelot2019mixmatch,berthelot2019remixmatch,sohn2020fixmatch,nassar2021all,zhang2021flexmatch,sosea2023marginmatch} combine these two techniques by using unsupervised data augmentation~\cite{xie2019unsupervised}. The main idea is to use the prediction of a weakly augmented image as the pseudo-label, and then enforce it to be consistent with the one with strong augmentation. For instance, UDA~\cite{xie2019unsupervised} and ReMixMatch~\cite{berthelot2019remixmatch} apply confidence-based strategies and sharpened soft pseudo labels to select sufficiently confident data for training, whereas FixMatch~\cite{sohn2020fixmatch} uses one-hot hard labels. More recently, SemCo~\cite{nassar2021all} successfully complements FixMatch by introducing semantic prior knowledge from language models~\cite{pennington2014glove}. FlexMatch~\cite{zhang2021flexmatch} applies curriculum learning to design a dynamic threshold function and select high confident pseudo labels according to learning status. MarginMatch~\cite{sosea2023marginmatch} utilizes margin-based strategies to select high-confidence pseudo labels. We adopt three SSL algorithms: SemCo, FlexMatch and MarginMatch for pre-training, and investigate the impact of meta-training on SSL models.

\section{Practical Semi-Supervised Meta-Training}\label{sec:new_dataset}
We first introduce the dataset settings of both common FSL and SSL, and then unify these two settings to define our practical semi-supervised meta-training setting. 

\noindent\textbf{Few-Shot Learning (FSL).}
An FSL dataset is usually split as base and novel class sets, where there is no overlap class between these two sets. The first one is used to train an FSL model, and the latter one is utilized to evaluate the recognition ability of the FSL model for novel objects. Formally, we denote the base set as $\mathcal{D}_\mathit{base}=\{\mathbf{x}_i,y_i\}_\mathit{i=1}^\mathit{N_{b}}$, where $\mathbf{x}_i\in \mathbb{R}^{D}$ is the sample (image), $y_i \in \mathcal{Y}_{b}=(y^1,y^2,\cdots,y^{\mathit{C_b}})$ is the class label, and $N_b$ and $\mathit{C_b}$ are the numbers of sample and class, respectively. 
Similarly, the novel set is denoted as $\mathcal{D}_\mathit{novel}=\{\mathbf{x}_i, y_i\}_\mathit{i=1}^\mathit{N_{n}}$, $y_i \in \mathcal{Y}_n=\{y^j\}_\mathit{j=C_b+1}^\mathit{C_b+C_n}$, respectively, where $\mathit{C_n}$ represents the novel class number. $\mathcal{Y}_{b}$ and $\mathcal{Y}_n$ are denoted as the base and novel label sets, respectively.  

Most FSL methods use a meta-learning manner (episodic sampling) to generate many few-shot tasks for training. Specifically, to generate a $M$-way $K$-shot task, we randomly select $M$ classes from the base set and then randomly choose $N_k+N_q$ samples from each selected class, where $MN_k$ and $MN_q$ samples are used to build the support set $\mathcal{S}$ and query set $\mathcal{Q}$, respectively. 
The goal is to find a model $f_\mathit{fsl}$ that classifies a query sample $\mathbf{x}_q$ in $\mathcal{Q}$ by $\hat{y}_q=f_\mathit{fsl}(\mathbf{x}_q;\theta_\mathit{fsl},\mathcal{S})\in \mathcal{Y}_s \subset \mathcal{Y}_{b}$, 
where $\mathcal{Y}_s$ is the class label set of $\mathcal{S}$, and $\theta_\mathit{fsl}$ is the parameter set of $f_\mathit{fsl}$. 
Formally, the model $f_\mathit{fsl}$ is learned to minimize the averaged loss over those generated tasks. The loss is defined: 
\begin{equation}
    \label{eq:fsl_loss}
    \mathit{L}_f=\sum\limits_{(\mathbf{x},y)\in\mathcal{Q}}l_f(\mathbf{x},y;\theta_\mathit{fsl},\mathcal{S}),
\end{equation}
where the loss $l$ measures the discrepancy between the prediction and the true label. 
Finally, we evaluate the model on thousands of few-shot tasks on $\mathcal{D}_\mathit{novel}$ and take the average classification accuracy as the final performance. 

\noindent\textbf{Semi-Supervised Learning (SSL).}
Different from FSL, SSL does not have a novel class set for evaluation or testing. The classes in the testing set of SSL are the same as the training set. Thus, in this setting, we do not need to use $\mathcal{D}_\mathit{novel}$. To evaluate the performance on both few-shot and semi-supervised tasks, we split the original base set $\mathcal{D}_\mathit{base}$ into three sets: $\mathcal{D}_{l}$, $\mathcal{D}_{u}$, and $\mathcal{D}_{t}$, which are respectively corresponding to the labeled, unlabeled, and testing set in the semi-supervised setting. Specifically, we randomly select $N_l$ (less than 20\%) and $N_t$ (near 20\%) samples from each class to build the labeled and testing set, respectively. The remaining samples in each class are used to build the unlabeled set. The goal of SSL is to learn a model $f_\mathit{ssl}$ that classifies the testing instance of base class $\mathbf{x}_{t}$ by $y_{t}=f_\mathit{ssl}(\mathbf{x}_{t};\theta_s)\in \mathcal{Y}_{b}$. Formally, we can learn the model by minimizing the following loss~\cite{yang2021survey},
\begin{equation}
    \sum\limits_{(\mathbf{x},y)\in\mathcal{D}_{l}}\mathit{l}_l(\mathbf{x},y;\theta_s)
    +\alpha\sum\limits_{\mathbf{x}\in\mathcal{D}_{u}}\mathit{l}_u(\mathbf{x};\theta_s)
    +\beta \sum\limits_{\mathbf{x}\in \bar{\mathcal{D}}_\mathit{train}}\mathit{r}(\mathbf{x};\theta_s),
    \label{eq:ssl_loss}   
\end{equation}
where $\mathit{l}_l$ is the per-example supervised loss, $\mathit{l}_u$ denotes the unsupervised loss, $\mathit{r}$ is the regularization, \emph{e.g.}, the consistency loss or a regularization term, $\mathcal{\Bar{D}}_\mathit{train}=\mathcal{D}_{l} \cup \mathcal{D}_{u}$.

\noindent\textbf{Semi-Supervised Meta-Training for Few-Shot Learning.}
Given the splitting sets, we can use $\mathcal{\Bar{D}}_\mathit{train}$ as the training (base) set in the newly proposed semi-supervised meta-training setting. The novel set is the same as the common few-shot setting to maintain the same evaluation for existing methods. In our experiments, we evaluate our model on the few-shot novel sets $\mathcal{D}_\mathit{novel}$, but also test it on semi-supervised testing set $\mathcal{D}_{t}$ to explore the impact of the meta-learning manner for common semi-supervised approaches.

Our new training setting is totally different to the previous SSFSL methods, including meta-training \cite{ren2018meta,li2019learning} and transfer-learning \cite{wang2020instance,rodriguez2020embedding} based algorithms.  
Firstly, in meta-training based works, there exists an additional unlabeled set $\mathcal{U}$ for each $M$-way $K$-shot task. $U$ unlabeled samples from each of $M$ classes are selected to build this unlabeled set. However, this selection strategy requires prior information for the labels of these 'unlabeled' samples, which is not reasonable during the training phase and can not utilize the real unlabeled samples. To remove the requirement of prior labels and make a  practical semi-supervised few-shot setting, we split the training set into labeled ($\mathcal{D}_{l}$) and unlabeled ($\mathcal{D}_{u}$) sets, and forbid the above selection strategy on the unlabeled set during the training phase. Secondly, the algorithms based on transfer-learning require all labeled data in the base class set for training, while our setting only needs a few labels. 

\begin{figure*}[!htb]
\centering
\includegraphics[width = 1\textwidth]{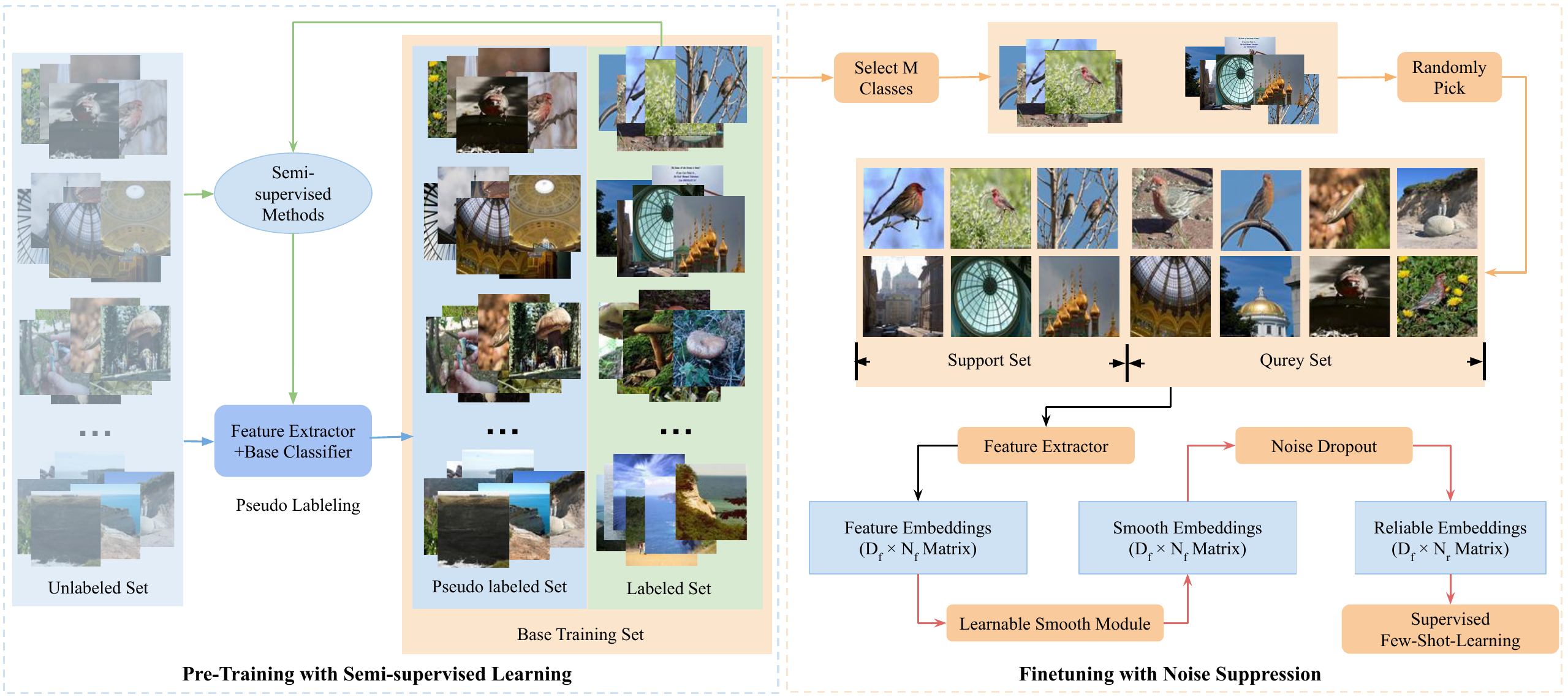} \\

\caption{
\textbf{The framework of our pseudo-labeling based meta-learning.} Our framework contains two stages: pre-training with semi-supervised learning and fine-tuning with noise suppression.
}\label{fig:plml}
\end{figure*}

\section{Pseudo-Labeling Based Meta-Learning}
Under the proposed semi-supervised few-shot setting, we firstly ignore the unlabeled set $\mathcal{D}_{u}$ and use the label set $\mathcal{D}_{l}$ to train existing few-shot models. We change the label number $N_l$ and observe the performance of three representative methods (Proto~\cite{snell2017prototypical}), T-EP, and S-EP~\cite{rodriguez2020embedding} on several few-shot tasks. As shown in Fig.~\ref{fig:p-degrade}, the evaluation scores of all models dramatically drop with the reduction of $N_l$. For example, when we use a very small number of labeled data \emph{i.e.} $N_l=20$ (4\% $|\mathcal{\Bar{D}}_\mathit{train}|$), the accuracy of S-EP model (with Conv4) on \textit{mini}ImageNet~\cite{ravi2016optimization} decreases sharply from 72.45\% to 55.64\% in 5-way 5-shot task. Can we use the unlabeled data to fill up the huge gap caused by the reduction of labeled data? To answer this question, we propose a novel semi-supervised few-shot training framework to make full use of the unlabeled data. Inspired by the recent two-stage method~\cite{rodriguez2020embedding}, we also decompose the training method as pre-training and finetuning stages. We illustrate our framework in Fig.~\ref{fig:plml}.

\subsection{Pre-Training with SSL}
The FSL model $f_\mathit{fsl}$ is often decomposed as a feature extractor $\phi$, which is used to obtain the discriminative feature of a sample $\mathbf{x}$, \emph{i.e.} $\mathbf{z} = \phi(\mathbf{x};\theta_{\phi})$, and a $M$-way classifier $c_{f}$, which is used to predict the class label according the support set $\mathcal{S}$, \emph{i.e.} $\mathbf{\hat{y}}=c_{f}(\mathbf{z};\theta_{f}, \mathcal{S})\in \mathbb{R}^{M}$, where $\mathbf{\hat{y}}$ is the predicting one-hot label of $\mathcal{Y}_s$. The aim of pre-training is to achieve a good initialization of $\phi$, which contains sufficient class prior knowledge. Thus, the previous methods \cite{wang2020instance,rodriguez2020embedding} add a new base classifier $c_b$ for the base set, \emph{i.e.} $\mathbf{y}=c_{b}(\mathbf{z};\theta_b)\in\mathbb{R}^{C_b}$, where $\mathbf{y}$ is the predicting one-hot label of $\mathcal{Y}_{b}$, in contrast to $c_{f}$ only focusing classes in the support set. Then $\phi$ and $c_b$ are trained via the supervised learning method. Surprisingly, this strategy obtains significant improvement in the few-shot task.  

Inspired by this, we also use $c_b$ for pre-training in our new semi-supervised FSL approach. Our goal is not only to train a good feature extractor $\phi$ but also to learn a discriminative base classifier $c_b$ to predict the base class labels of the unlabeled samples. These predicted labels are used as pseudo labels in the finetuning stage. We wish the pre-training method could improve the discriminate ability of both $\phi$ and $c_b$. This goal is the same as the common semi-supervised learning, which aims to learn a model that classifies the instances with the base class labels ($\mathcal{Y}_{b}$). 
According to Eq.~\ref{eq:ssl_loss}, 
We set $\{\theta_{\phi},\theta_b\}\subseteq\theta_s$. Then, we can directly adopt the existing semi-supervised approaches to pre-train our feature extractor $\phi$ and classifier $c_b$. 

\subsection{Finetuning with Noise Suppression}\label{sec:fns}
To utilize the unlabeled data, we firstly use the pre-trained model to generate pseudo-labels as follows,
\begin{equation}
    \label{eq:pseudo-labels}
    \Tilde{y}_i=\pi(\mathbf{x}_i)=\mathcal{Y}_{b}[\arg\max_{j} c_b(\phi(\mathbf{x}_i))[j]], \mathbf{x}_i\in\mathcal{D}_u.
\end{equation}
We denote $\mathcal{D}_{p}=\{(\mathbf{x}_i,\Tilde{y}_i) | \mathbf{x}_i \in \mathcal{D}_u\}$ as the pseudo-labeling dataset.
Combined with the labeled set, we obtain the new training set $\mathcal{\Tilde{D}}_\mathit{train}=\mathcal{D}_l\cup\mathcal{D}_p$, which might contain noisy labels. To reduce the negative impact of noisy labels, we propose 
a novel noise-suppressed finetuning framework based on label consistency with similar features. Our core idea is inspired by the assumption that samples with similar features have the same label with a high possibility. Accordingly, we design a learnable smoothing module for feature adjustment and a noise dropout strategy for finetuning. 

\noindent\textbf{Learnable Smoothing Module.} For each few-shot task, we aim to adjust the noisy samples by utilizing all samples in the few-shot task, in order to smooth their features to be more similar to the samples with correct labels. Specifically, first, we adopt a Transformer encoder layer of depth one $\mathcal{T}$ \cite{vaswani2017attention}, so as to smooth feature embeddings through self attention between all samples of a few-shot task. Specifically, denote the feature matrix as $\mathbf{Z}=[\mathbf{z}_1;\mathbf{z}_2;\cdots;\mathbf{z}_{N_f}]\in \mathbb{R}^{N_f \times D_f}$, where $D_f$ is the feature dimension, $N_f$=$| \mathcal{\Tilde{Q}}\cup\mathcal{\Tilde{S}}|$, and $\mathcal{\Tilde{S}}$ and $\mathcal{\Tilde{Q}}$ respectively represent the support and query sets produced from $\mathcal{\Tilde{D}_\mathit{train}}$.
Then the transformer operation is formulated as $\mathbf{Z}_t=\mathcal{T}(\mathbf{Z};\theta_t)$, where $N_f$ in $\mathbf{Z}$ is interpreted as the sequence length or the number of words in Transformers, so that the self attention mechanism in Transformers can be considered as a feature propagation and smoothing layer.
Furthermore, to make this module learn more from similar samples, we add another feature (embedding) propagation mechanism \cite{zhou2003learning,rodriguez2020embedding} to guide the learning procedure and adjust noisy features to be more smooth. Specifically, firstly we compute the adjacency matrix as $\mathbf{A}_{ij}=\exp{(-d_{ij}^2/\sigma^2)}$, if $i\neq j$, while $\mathbf{A}_{ii}=0$, where $d_{ij}^2=\Arrowvert \mathbf{z}_i^t-\mathbf{z}_j^t \Arrowvert^2_2$, $\sigma^2=\mathit{Var}(d_{ij}^2)$, $\forall \mathbf{z}_i^t,\mathbf{z}_j^t \in \mathbf{Z}_t$.
Given the adjacency matrix, we obtain the propagator matrix $\mathbf{P}=(\mathbf{I}-\alpha_p\mathbf{L})^{-1}$, where $\mathbf{L}=\mathbf{D}^{-\frac{1}{2}}\mathbf{A}\mathbf{D}^{-\frac{1}{2}}$ is the Laplacian matrix \cite{zhou2003learning}, $\mathbf{D}_{ii}=\sum_j\mathbf{A}_{ij}$, 
$\alpha_p=0.2$ is a scaling factor, and $I$ is the identity matrix. Then, the features are adjusted as follows, $\mathbf{Z}_p=\mathbf{P}\mathbf{Z}_t$.
The learnable smoothing module is: 
\begin{equation}\label{eq:lsm}
    \mathbf{z}^p_i = m_\mathit{s}(\mathbf{z}_i;\theta_t),\forall\mathbf{z}_i^p\in\mathbf{Z}_p.
\end{equation}

\noindent\textbf{Noise Dropout Strategy.}
To further suppress the noise, we drop out some labels according to the assumption of label consistency among similar samples. For $i$-th class set in an $M$-way $K$-shot task, we randomly select a sample as an anchor and denote its feature as $\mathbf{z}_i^a\in\mathbf{Z}$. We compute the distances between the anchor and the other samples, $d_{ij}^a=\Arrowvert \mathbf{z}_i^a-\mathbf{z}_j \Arrowvert^2_2, \forall \mathbf{z}_j\in\mathbf{Z},\hat{y}_j=i$. Then we sort these samples in ascending order of distance and select the top $N_k$ samples (including the anchor) to rebuild the support set, which will contain less noise. Since the samples with high distances are more likely to have inconsistency labels with the anchor, we drop out the last $\lfloor\alpha_dN_q\rfloor$ samples and select the remaining samples to rebuild the query set, where $\alpha_d$ is the dropout rate (experimentally set as 0.1), and $N_q$ is the query shot. Finally, we obtain more clean support set $\mathcal{\Tilde{S}}_r$ and query set $\mathcal{\Tilde{Q}}_r$.

\noindent\textbf{Training Loss.}
Our learnable smoothing module and noise dropout strategy are model-agnostic, and we can incorporate them with most few-shot models. Combined Eq.~\ref{eq:lsm} with Eq.~\ref{eq:fsl_loss}, we obtain the general formulation of loss as follows,
\begin{equation}
    \label{eq:l_f}
    L_f = \sum\limits_{(\mathbf{x},\Tilde{y})\in\mathcal{\Tilde{Q}}_r}l_f(\mathbf{x},\Tilde{y};\theta_\phi,\theta_t,\theta_f,\mathcal{\Tilde{S}}_r),
\end{equation}
where $l_f(\mathbf{x}_i, \Tilde{y}_i)=H(c_f(m_s(\phi(\mathbf{x}_i))),\mathbf{\hat{y}}_i)$, $H$ represents the cross-entropy loss and $\mathbf{\hat{y}}_i\in \mathbb{R}^{M}$ is the one-hot label of $M$ classes. 
Besides, we finetune the classifier $c_b$ to preserve the discriminative ability on the basic classes by
\begin{equation}
    \label{eq:l_l}
L_l=\sum\limits_{(\mathbf{x},\Tilde{y})\in\mathcal{\Tilde{Q}}\cup\mathcal{\Tilde{S}}} l_l(\mathbf{x},\Tilde{y};\theta_\phi,\theta_b),
\end{equation}
where $l_l(\mathbf{x}_i, \Tilde{y}_i) =H(c_b(\phi(\mathbf{x}_i)),\mathbf{y}_i)$, and $\mathbf{y}_i\in \mathbb{R}^{C_b}$ is the one-hot label. 
The final loss is $L_{\mathit{ft}}=L_f+\gamma L_l$, 
where the trade-off parameter $\gamma$ is empirically set as 0.1. 

\tr{We present the pseudo-code for our proposed framework in Algorithm~\ref{algo-PLML}. This pseudo-code outlines the key steps and components, providing a clear and concise guide for applying our method. The specific code can be found at the GitHub address mentioned in the abstract.}

\begin{algorithm}[htb]
\caption{\tr{Pseudo-Labeling based Meta-Learning}}
    \textbf{Input:} Labeled Set $D_l$, Unlabeled Set $D_u$, Epoch Num $E$, Batch Num $B$, Task Num $T$, Way Num $M$, Shot Num $N_k$, Query Num $N_q$, Feature Dimension $D_f$, Drop Rate $\alpha_d$, Trade-off parameter $\lambda$ \\
    \textbf{Output:} $\theta_{\phi}$, $\theta_b$, $\theta_t$, $\theta_f$
    
    \begin{algorithmic}[1]
    \STATE \textbf{Pre-training with SSL:} Obtain a feature extractor $\phi$ and a discriminative base classifier $c_b$
    \STATE $\tilde{D}_{train} \leftarrow D_l$
    \FOR {each $x_i$ in $D_u$}
        \STATE Calculate pseudo-labels $\tilde{y}_i$ using Eq.~\ref{eq:pseudo-labels}
        \STATE $\tilde{D}_{train}.add((x_i, \tilde{y}_i))$
    \ENDFOR
    \FOR {epoch $e$ in range($E$)}
        \FOR {batch $b$ in range($B$)}
            \FOR {task $t$ in range($T$)}
                \STATE Sample a few-shot task with support set $\tilde{S}$ and query set $\tilde{Q}$ from $\tilde{D}_{train}$
                \STATE $N_f \leftarrow |\tilde{Q} \cup \tilde{S}|$
                \STATE $Z \leftarrow \text{zero}(N_f, D_f)$
                \FOR {each $x_i$ in $\tilde{Q} \cup \tilde{S}$}
                    \STATE $z_i \leftarrow \phi(x_i)$
                \ENDFOR
                \STATE $\mathbf{Z}_t \leftarrow \mathcal{T}(\mathbf{Z})$
                \STATE $Z_p \leftarrow P Z_t$ using Eq.~\ref{eq:lsm}
                \FOR {i in range($M$)}
                    \STATE Randomly sample $z^a_i$ in $Z$ as anchor
                    \FOR {each $z_j$ in $Z$ with $y_j = i$}
                        \STATE Calculate distance $d^a_{ij} = \|z^a_i - z_j\|^2$
                    \ENDFOR
                    \STATE $\tilde{Q}_{ir} \cup \tilde{S}_{ir} \leftarrow (Q_i \cup S_i).sort(d^a_{ij}).remove(\lfloor \alpha_d \cdot N_q \rfloor \text{ largest samples})$
                \ENDFOR
                \STATE Calculate loss $L_{ft} \leftarrow L_f + \lambda L_l$, where $L_f$ is defined in Eq.~\ref{eq:l_f}, $L_l$ is defined in Eq.~\ref{eq:l_l}
            \ENDFOR
        \ENDFOR
    \ENDFOR
    
    \end{algorithmic}
\label{algo-PLML}
\end{algorithm}

\section{Experiments}\label{sec:exp}

\subsection{Experimental Setting}
\noindent\textbf{Datasets.} 
We evaluate our approach in two standard datasets: \textit{mini}ImageNet~\cite{ravi2016optimization}, and  \textit{tiered}ImageNet~\cite{ren2018meta}. 
\textit{mini}ImageNet is a subset of ImageNet~\cite{deng2009imagenet} composed of 100 classes with 600 images per class. Classes are divided into 64
base, 16 validation, and 20 novel categories.
\textit{tiered}ImageNet is a more challenging and larger dataset than \textit{mini}ImageNet, 
where classes are split into 20 base (351 classes), 6 for validation (97 classes), and 8 novel (160 classes) super categories, which are selected from supersets of the wordnet hierarchy. Each class contains nearly 900--1200 images.
To adjust the above datasets for our new SSFSL setting, we set different $N_l$ and $N_t$ proposed in \S\ref{sec:new_dataset} to split labeled, unlabeled, and testing sets for different datasets, according to the number of images per class.
Specifically, for \textit{mini}ImageNet, we set $N_l=[20, 50, 100]$ and $N_t=100$. We increase the numbers of \textit{tiered}ImageNet, \emph{i.e.}, $N_l=[20, 100, 200]$ and $N_t=200$. 
Notice that our results with all base training data are lower than the ones in the original papers since we use fewer labels.

\noindent\textbf{Training and Inference Details.}
We use two common feature extractors (backbones): (i) Conv4: a 4-layer convolutional network \cite{snell2017prototypical} with 64 channels per layer, and (ii) ResNet12: a 12-layer residual network \cite{oreshkin2018tadam}. For all datasets, we resize images to 84$\times$84. 
In the pre-training stage, we use the same strategy in EP~\cite{rodriguez2020embedding} as the Base training method and train the SSL models (SemCo~\cite{nassar2021all}, FlexMatch~\cite{zhang2021flexmatch} and {MarginMatch~\cite{sosea2023marginmatch}}) by using the default hyperparameter setting in official codes. 
In the episodic finetuning stage, for \textit{mini}ImageNet and \textit{tiered}ImageNet, 5 classes are randomly sampled per episode, where in each class we select 5 and 15 instances for the support and query set, respectively. 
We optimize the models by using SGD with an initial learning rate of 0.01 for 200 epochs. When the validation loss had not decreased for 10 epochs, we reduce the learning rate by a factor of 10. 
We use 15 instances in the query set of 1- or 5-shot tasks for inference.
We adopt 100 unlabeled data per class (same as EP) for semi-supervised inference.

\begin{table*}[htb!]
\centering
\begin{tabular}{lll|cccc|cccc}
\hline
Methods                  & Pre-training    & Backbone & \multicolumn{4}{c}{5-way-1-shot} & \multicolumn{4}{|c}{5-way-5-shot} \\
\hline
\multicolumn{3}{r|}{Labels Per Class $\longrightarrow$}                  & 20     & 50     & 100    & Full   & 20     & 50     & 100    & Full   \\
\hline
SPN~\cite{ren2018meta}                 & N/A             & Conv4    & 28.03  & 34.24  & 40.44  & N/A   & 38.37  & 47.94  & 54.78  & N/A   \\
M-PL~\cite{li2022platinum}                    & N/A             & Conv4    & 40.76  & 44.72  & 45.85  & N/A   & 54.93  & 58.51  & 61.37  & N/A   \\
EP~\cite{rodriguez2020embedding}                      & Supervised      & Conv4    & 45.06  & 51.91  & 55.33  &\bf 60.33 & 55.64  & 64.08  & 66.27  &\bf 72.45 \\
MFC~\cite{ling2022semi}                       & Supervised      & Conv4    & 42.91  & 48.14  & 51.41  & 55.49 & 50.94  & 57.10  & 60.90  & 64.92 \\
SemCo-{\ourmodel}-EP (ours)     & Semi-supervised & Conv4    &\bf 57.33 &\bf 58.78 &\bf \red{62.67} & N/A  &\bf 69.19 &\bf 70.24 &\bf \red{73.45}  & N/A   \\
FlexMatch-{\ourmodel}-EP (ours) & Semi-supervised & Conv4    & 56.46 & 56.04 & 59.81 & N/A  & 68.36 & 68.14 & 70.77  & N/A   \\
{MarginMatch-{\ourmodel}-EP (ours)} & Semi-supervised & Conv4    & 55.13 & 56.96 & 60.13 & N/A  & 67.58 & 69.15 & 71.25  & N/A   \\
\hline
\hline
EP~\cite{rodriguez2020embedding}                       & Supervised      & ResNet12 & 49.29  & 58.05  & 62.63  & 73.15 & 62.44  & 70.23  & 74.13  & 82.39 \\
MFC~\cite{ling2022semi}                      & Supervised      & ResNet12 & 36.91  & 48.73  & 57.14  &\bf 74.06 & 48.31  & 62.15  & 69.56  &\bf 82.40 \\
SemCo-{\ourmodel}-EP (ours)     & Semi-supervised & ResNet12 & \bf 71.61 &\bf 72.84 &\bf 72.97 & N/A   & 81.51 &\bf \red{82.92} &\bf \red{83.70} & N/A   \\
FlexMatch-{\ourmodel}-EP (ours) & Semi-supervised & ResNet12 & 70.72  & 72.18  & 72.18  & N/A   &\bf 81.53  & 81.71  & 81.72  & N/A  \\
{MarginMatch-{\ourmodel}-EP (ours)} & Semi-supervised & ResNet12 & 65.90  & 69.90  & 70.50  & N/A   &\bf 76.25  & 78.93  & 80.04  & N/A  \\
\hline
\end{tabular}
\caption{
\textbf{Accuracy (\%) of $M$-way $K$-shot tasks with semi-supervised inferences on \textit{mini}ImageNet.} We show our results based on the FSL method of EP~\cite{rodriguez2020embedding}, and three SSL approaches: SemCo~\cite{nassar2021all}, FlexMatch~\cite{zhang2021flexmatch} and MarginMatch~\cite{sosea2023marginmatch} for pre-training. The best values are in bold. Red values indicate our {\ourmodel} outperforms the models trained with full labels. Notice that SPN and M-PL only use Conv4 for evaluation. Thus, we only provide the results of Conv4.
}
\label{tab:semi-fsl}
\end{table*}

\subsection{Previous Semi-supervised FSL methods with Fewer Labels}
{To investigate the effect of the number of training labels for previous Semi-supervised FSL methods, we select two kinds of mainstream algorithms including semi-supervised meta-training based and supervised pre-training based methods, as the representation for comparison. The former methods, such as SPN~\cite{ren2018meta} and M-PL~\cite{li2022platinum}, split the base training set as labeled and unlabeled sets, and only apply semi-supervised meta-training to learn FSL models. The latter methods use all labels in the base training set for pre-training by simple supervised learning to obtain discriminative feature extractors. 
}

{
As shown in Table~\ref{tab:semi-fsl}, we evaluate all approaches on the novel class set of \textit{mini}ImageNet regarding 5-way 1-shot and 5-shot tasks. 
All previous methods suffer from large performance degradation with reduced labeled data. The pioneering work SPN requires a large number of labels for training. In its original setting, the labeled set contains 40\% labeled samples of the base training set. Thus, when we reduce the labels, its performance dramatically drops. M-PL achieves more stable performance with the reduction of labels, since it uses submodular mutual information to select reliable unlabeled set for meta-training. However, these two methods still have a large gap with EP and MFC, because they lack of pre-training phase for better feature representation. EP and MFC also have large performance degradation with a few training labels. For example, on the setting of 20 labels per class, EP suffers from huge performance degradation on four tasks by decrements of 15.27\% (1 shot with Conv4), 16.80\% (5 shot with Conv4), and 23.86\% (1 shot with ResNet12), and obtains a similar reduction of 19.95\% in the last task. MFC also has similar dramatical decrements on four FSL tasks. 
}

{In summary, some performance gaps still exist between semi-supervised meta-training based methods and pre-training based semi-supervised FSL methods. Most semi-supervised models still require abundant labels to achieve the expected performance. These results inspired us to design a new semi-supervised FSL model to effectively utilize the labels. }

\begin{table*}[htb!]
\centering
\begin{tabular}{llll|cccc|cccc}
\hline
Pre-training  & Finetuning & Inference & Backbone & \multicolumn{4}{c}{5-way 1-shot} & \multicolumn{4}{|c}{5-way 5-shot} \\
\hline
\multicolumn{4}{r|}{Labels Per Class $\longrightarrow$}        & 20     & 50     & 100    & Full   & 20     & 50     & 100    & Full   \\
\hline
Base      & N/A      & \textit{in}-Proto     & Conv4    & 39.86  & 43.37  & 45.53  & 48.64 & 55.09  & 60.26  & 63.22  & 67.45 \\
Base      & Proto    & \textit{in}-Proto     & Conv4    & 38.58  & 43.37  & 44.89  &\bf 49.77 & 54.29  & 59.62  & 62.36  &\bf 68.58 \\
SemCo     & N/A      & \textit{in}-Proto     & Conv4    & 42.68  & 44.98  & 47.01  & N/A   & 61.71  & 63.55  & 65.71  & N/A   \\
SemCo     &\ourmodel-Proto (ours)    & \textit{in}-Proto     & Conv4    &\bf 45.24  &\bf 45.13  &\bf 47.57  & N/A   &\bf 64.94  &\bf 66.16  &\bf 68.32  & N/A   \\
FlexMatch & N/A      & \textit{in}-Proto     & Conv4    & 40.57  & 42.35  & 44.91  & N/A   & 59.01  & 61.69  & 62.92  & N/A   \\
FlexMatch &\ourmodel-Proto (ours)    & \textit{in}-Proto     & Conv4    & 42.55  & 43.75  & 46.63  & N/A   & 62.83  & 63.62  & 67.01  & N/A   \\
{MarginMatch} & N/A      & \textit{in}-Proto     & Conv4    & 41.46    & 41.10  & 44.23  & N/A   & 58.93  & 60.28  & 62.64  & N/A   \\
{MarginMatch} &\ourmodel-Proto (ours)    & \textit{in}-Proto     & Conv4    & 42.86  & 44.47  & 46.93  & N/A   & 62.87  & 64.49  & 66.75  & N/A   \\
\hline
Base      & N/A      & \textit{trans}-EP  & Conv4    & 43.33  & 47.01  & 49.54  & 53.06 & 55.92  & 60.36  & 63.74  & 67.23 \\
Base      & EP       & \textit{trans}-EP  & Conv4    & 42.31  & 46.78  & 49.01  &\bf 54.99 & 54.79  & 61.07  & 63.60  &\bf 68.92 \\
SemCo     & N/A      & \textit{trans}-EP  & Conv4    & 47.56  & 49.44  & 52.10  & N/A   & 61.39  & 64.18  & 66.57  & N/A   \\
SemCo     & \ourmodel-EP (ours)       & \textit{trans}-EP  & Conv4    &\bf52.59 &\bf 54.03 &\bf \red{56.23} & N/A& \bf 66.30 &\bf 67.92 &\bf \red{70.05} &  N/A \\
FlexMatch & N/A      & \textit{trans}-EP  & Conv4    & 45.35  & 47.77  & 49.57  & N/A   & 58.75  & 61.76  & 64.02  & N/A   \\
FlexMatch & \ourmodel-EP (ours)       & \textit{trans}-EP  & Conv4    & 50.59 & 52.22 & \red{55.14} & N/A& 64.47 & 65.29 & 68.78  & N/A   \\
{MarginMatch} & N/A      & \textit{trans}-EP  & Conv4    & 45.17  & 48.32  & 49.83  & N/A   & 59.48  & 60.99  & 63.13  & N/A   \\
{MarginMatch} & \ourmodel-EP (ours)       & \textit{trans}-EP  & Conv4    & 49.93 & 52.13 & 54.39 & N/A& 63.81 & 65.88 & 67.85  & N/A   \\

\hline

\hline
Base      & N/A      & \textit{in}-Proto                & ResNet12 & 37.06  & 41.41  & 45.37  & 51.05 & 55.95  & 63.50  & 68.70  & 75.97 \\
Base      & Proto    & \textit{in}-Proto                & ResNet12 & 34.97  & 41.13  & 43.43  &\bf 54.20 & 54.71  & 62.95  & 66.32  &\bf 76.21 \\
SemCo     & N/A      & \textit{in}-Proto                & ResNet12 & 46.52  & 46.82  & 47.35  & N/A   & 72.27  & 73.38  & 73.96  & N/A   \\
SemCo     & \ourmodel-Proto (ours) & \textit{in}-Proto  &ResNet12  & 49.74  & 49.21  & 51.15  & N/A   &\bf \red{76.35}  &\bf \red{76.25}  &\bf \red{76.50}  & N/A   \\
FlexMatch & N/A      & \textit{in}-Proto                & ResNet12 & 45.40  & 46.30  & 45.52  & N/A   & 70.55  & 71.85  & 70.79  & N/A   \\
FlexMatch & \ourmodel-Proto (ours)  & \textit{in}-Proto & ResNet12 & 47.03  & 47.54  & 48.09  & N/A   & 72.46  & 73.20  & 75.03  & N/A   \\
{MarginMatch} & N/A    & \textit{in}-Proto                & ResNet12 & 45.84  & 45.71  & 44.36  & N/A   & 68.31  & 70.44  & 69.08  & N/A   \\
{MarginMatch} & \ourmodel-Proto (ours) &\textit{in}-Proto & ResNet12 & \bf 50.15  & \bf 50.84  & \bf 51.66  & N/A   & 71.96  & 73.22  & 75.09  & N/A   \\
\hline
Base      & N/A  & \textit{trans}-EP & ResNet12 & 41.54 & 48.09 & 52.43 & 60.05 & 54.93 & 63.02 & 68.65 & 76.28 \\
Base      & EP    & \textit{trans}-EP & ResNet12 & 44.77 & 50.63 & 54.93 &\bf 65.75 & 58.13 & 65.72 & 69.72 &\bf 79.16 \\
SemCo     & N/A  & \textit{trans}-EP & ResNet12 & 60.34 & 61.57 & 62.32 & N/A& 75.10 & 76.11 & 76.47 & N/A\\
SemCo     & \ourmodel-EP (ours)    & \textit{trans}-EP & ResNet12 &\bf 64.45 &\bf 64.78 & 64.40 & N/A&\bf 77.67 & 78.00 &\bf 78.84 & N/A\\
FlexMatch & N/A  & \textit{trans}-EP & ResNet12 & 52.50 & 54.20 & 52.75 & N/A& 69.53 & 71.13 & 69.70 & N/A\\
FlexMatch & \ourmodel-EP (ours)    & \textit{trans}-EP & ResNet12 & 63.01 & 64.39 &\bf 64.52 & N/A& 76.96 &\bf 78.29 & 78.65 & N/A\\
{MarginMatch} & N/A  & \textit{trans}-EP & ResNet12 & 55.86 & 57.89 & 58.17 & N/A& 70.19 & 73.10 & 73.28 & N/A\\
{MarginMatch} & \ourmodel-EP (ours)    & \textit{trans}-EP & ResNet12 & 59.55 & 61.26 & 62.86 & N/A& 73.84 & 75.91 & 77.13 & N/A\\
\hline
\end{tabular}
\caption{
\textbf{Accuracy (\%) of $M$-way $K$-shot tasks with inductive (\textit{in}), transductive (\textit{trans}) and semi-supervised (\textit{semi}) inferences on \textit{mini}ImageNet.} We show the results of two FSL methods: Proto~\cite{snell2017prototypical} and EP~\cite{rodriguez2020embedding}, based on a Base pre-training method and two SSL approaches: SemCo~\cite{nassar2021all} and FlexMatch~\cite{zhang2021flexmatch}. The best values are in bold. Red values indicate our {\ourmodel} outperforms the models trained with full labels.
}
\label{tab:mini}
\end{table*}

\subsection{Inductive and Transductive Methods with Fewer Labels}
{To investigate the effect of the number of training labels in the other two kinds of FSL models, we select Proto~\cite{snell2017prototypical} as the representation for inductive FSL methods, and choose EP~\cite{rodriguez2020embedding} for transductive FSL algorithms. We first use the pre-training method of EP (\emph{e.g.} Base in Table~\ref{tab:mini}) to train models on the subsets of \textit{mini}ImageNet with different numbers of labeled data. Then we finetune these models by the episodic training (meta-learning) manners of Proto or EP on the same subsets. Finally, we evaluate these models on the novel class set by using commonly used two inferences in terms of 5-way 1-shot and 5-shot tasks. }
{To further investigate the impact of pre-training, we remove the finetuning phase and directly evaluate the pre-trained model, such as Base+N/A in Table~\ref{tab:mini}. 

When we reduce the number of labels, the finetuning phase can not obtain improvement in most tasks. Only Base+EP with ResNet12 achieves better performance (1/4) than the pre-training model. This indicates that under the situation of a few base labels, existing methods are not stable. Thus, we need to seek new FSL finetuning methods for this new situation. 
Moreover, as shown in Table~\ref{tab:mini}, all models suffer from huge performance degradation with the reduction of labeled data. For example, under the setting of 20 labels per class, the accuracy of inductive Proto with Conv4 (Base+Proto) declines by 11.19\% and 14.29\% in terms of 1-shot and 5-shot tasks, respectively. For the models of ResNet12 (Base+EP), amounts of reduction are larger and reach 19.23\% and 21.49\%. The performance of transductive EP with Conv4 drops sharply by 12.67\% and 14.14\% in two tasks. Similarly, the ResNet12 version losses more accuracy, declining by 20.98\% and 21.03\%. 
The above results indicate that current FSL models still rely heavily on mass labels in the base training set, which limits them from being used for practical applications.
}

\begin{table*}[!htp]
\centering
\begin{tabular}{llll|cccc|cccc}
\hline
Pre-training  & Finetuning & Inference & Backbone & \multicolumn{4}{c}{5-way 1-shot} & \multicolumn{4}{|c}{5-way 5-shot} \\
\hline
\multicolumn{4}{c|}{Labels Per Class $\longrightarrow$}        & 20     & 100     & 200    & All   & 20     & 100     & 200    & All   \\
\hline
Base      & Proto & \textit{in}-Proto    & Conv4    & \bf 45.61 & 50.80 & 51.07 &\bf 52.50 & 62.43 & 68.48 & 70.12 &\bf 71.50 \\
SemCo     & \ourmodel-Proto & \textit{in}-Proto    & Conv4    & \rec{45.17} &\bf 51.86 &\bf 52.04 & -  & \rec{64.25} &\bf \red{72.43} &\bf \red{72.64} & -  \\
FlexMatch & \ourmodel-Proto & \textit{in}-Proto    & Conv4    & 44.94 & 49.33 & 50.60 & -  & \bf 64.80 & 70.54 & 71.32 & -  \\
\hline
Base      & EP    & \textit{trans}-EP & Conv4    & 49.33 & 55.53 & 56.08 &\bf 58.10 & 62.71 & 68.67 & 70.01 &\bf 71.44 \\
SemCo     & \ourmodel-EP    & \textit{trans}-EP & Conv4    & \rec{52.45} &\bf \red{60.52} &\bf \red{61.09} & -  & \rec{65.49} &\bf \red{73.90} &\bf \red{74.29} & -  \\
FlexMatch & \ourmodel-EP    & \textit{trans}-EP & Conv4    & \bf 54.41 & \red{58.88} & \red{60.56} & -  & \bf 68.43 & \red{72.71} & \red{73.54} & -  \\
\hline
Base      & EP    & \textit{semi}-EP  & Conv4    & 54.02 & 61.19 & 61.96 &\bf 63.28 & 65.03 & 70.87  & 71.98 &\bf 74.14 \\
SemCo     & \ourmodel-EP    & \textit{semi}-EP  & Conv4    & \rec{56.85} &\bf \red{67.71} &\bf \red{68.20} & -  & \rec{68.20} &\bf \red{75.74} &\bf \red{76.66} & -  \\
FlexMatch & \ourmodel-EP    & \textit{semi}-EP  & Conv4    & \bf 60.50 & \red{65.93} & \red{66.37} & -  & \bf 71.38 & \red{74.85} & \red{75.28} & -  \\
\hline
\hline
Base      & Proto & \textit{in}-Proto    & ResNet12 & 44.59 & \bf 56.03 &\bf 59.73 & \bf 64.24 & 68.91 & 78.27 & 80.90 &\bf 83.66 \\
SemCo     & \ourmodel-Proto & \textit{in}-Proto    & ResNet12 & 49.55 & 55.35 & 57.41 & -  & 74.79 &\bf 80.48 &\bf 82.17 & -  \\
FlexMatch & \ourmodel-Proto & \textit{in}-Proto    & ResNet12 &\bf 51.60 & 54.12 & 58.25 & -  &\bf 75.97 & 79.66 & 81.61 & -  \\
\hline
Base      & EP    & \textit{trans}-EP & ResNet12 & 55.68 & 66.67 & 68.40 &\bf 72.66 & 71.19 & 80.79 & 80.96 &\bf 84.59 \\
SemCo     & \ourmodel-EP    & \textit{trans}-EP & ResNet12 &\bf 68.08 &\bf 70.69 &\bf \red{73.51} & -  &\bf 81.08 &\bf 83.09 &\bf \red{84.75} & -  \\
FlexMatch & \ourmodel-EP    & \textit{trans}-EP & ResNet12 & 66.65 & 69.98 & 71.81 & -  & 79.29 & 82.53 & 83.66 & -  \\
\hline
Base      & EP    & \textit{semi}-EP  & ResNet12 & 65.15 & 74.62 & 75.52 &\bf 79.97 & 75.60 & 83.10 & 83.54 &\bf 86.61 \\
SemCo     & \ourmodel-EP    & \textit{semi}-EP  & ResNet12 & \bf 75.13 & 76.87 &\bf 79.62 & -  &\bf 82.42 &\bf 85.07 &\bf \red{86.69} & -  \\
FlexMatch & \ourmodel-EP    & \textit{semi}-EP  & ResNet12 & 73.93 &\bf 77.02 & 79.12 & -  & 81.55 & 84.16 & 85.88 & -  \\
\hline

\end{tabular}
\caption{
\textbf{Accuracy (\%) of $M$-way $K$-shot tasks with inductive (\textit{in}), transductive (\textit{trans}) and semi-supervised (\textit{semi}) inference on \textit{tiered}ImageNet~\cite{ren2018meta}.} We show the results of two FSL methods: Proto~\cite{snell2017prototypical} and EP~\cite{rodriguez2020embedding}, based on a Base pre-training method and two SSL approaches SemCo~\cite{nassar2021all} and FlexMatch~\cite{zhang2021flexmatch}. \ourmodel{} represents our approach. The best values are in bold. Red values indicate our models outperform the original models trained with all labels.
}
\label{tab:tiered}
\end{table*}

\subsection{Benefits of Our \ourmodel}
To evaluate the effectiveness of our \ourmodel{}, we select SemCo~\cite{nassar2021all} as the pre-training method and then use our \ourmodel{}-Proto and \ourmodel{}-EP methods (based on Proto and EP) for finetuning. As shown in Tables~\ref{tab:semi-fsl} and \ref{tab:mini}, all models with our \ourmodel{} (SemCo+\ourmodel-*) achieve significant performance improvement in all evaluation tasks with three inference manners and fewer labels, compared with their baselines (Base+Proto or EP), especially on the dataset with the least labels (20). For example, under the setting of 20 labels per class, our models with ResNet12 obtain huge gains of 14.77\%--22.31\% in terms of 1-shot tasks with three inference manners. In the 5-shot task, the proposed models outperform baselines by large gains of 19.09\%--21.63\%. For the Conv4 backbone, our models also achieve significant gains of 6.66\%--12.27\% in the 1-shot task and 10.65\%--13.54\% in the 5-shot task. The above performance increments clearly demonstrate the benefits of our approach for FSL with a few training labels. In several settings, our models with fewer labels achieve superior performance than the models with full labels.
This suggests that the gap between full-supervised and semi-supervised training has been largely closed in FSL tasks.

\subsection{Different SSL Methods}
{To evaluate the generalization of our \ourmodel{}, we use the other SSL method, FlexMatch~\cite{zhang2021flexmatch} and MarginMatch~\cite{sosea2023marginmatch}, as the pre-training method. As shown in Table~\ref{tab:semi-fsl} and \ref{tab:mini}, our models based on FlexMatch and MarginMatch~\cite{sosea2023marginmatch} also obtain significant improvement over their baselines in most settings. This indicates that our \ourmodel{} is robust for different SSL pre-training methods. The overall performance of our FlexMatch-based and MarginMatch-based models is slightly lower than the corresponding SemCo-based models since SemCo introduces the semantic embedding features to provide better prior knowledge.} This suggests that our approach can transfer the advantages of SSL methods to the semi-supervised training of FSL, and bridge two topics.
Furthermore, we directly evaluate these pre-training methods on the novel set to explore the effectiveness of SSL models. {Compared with baselines (Base), both SemCo-, FlexMatch- and MarginMatch-based models achieve large gains in most settings.} However, there still exist gaps in the models with full labels, especially for the ResNet12-based models. Incorporated our \ourmodel{}, the gaps have been largely closed in most tasks. For example, our SemCo-based models with ResNet12 obtain close or superior performance in 5-way 5-shot tasks, compared with models trained by full labels.

\subsection{Our \ourmodel{} for Different Datasets}

To further demonstrate the generalization, we evaluate our approach in the \textit{tiered}ImageNet dataset. 
We show the results of 1- and 5-shot tasks in Table~\ref{tab:tiered}. 
Our \ourmodel{} facilitates both SemCo- and FlexMatch-based models to achieve significant improvement in most tasks and approach the baselines with full labels. \rec{Especially, in the 5-way 1-shot tasks on 20 labels subset, the highest gain reaches 12.4\% (SemCo+\textit{trans}-EP+ResNet12). Furthermore, our models outperform the baselines with full labels in 21 tasks (highlighted by red).} This demonstrates our generalization ability for the large-scale dataset. 

\subsection{Different SS-Meta-Training Setting}
To investigate the impact of class-prior selection strategy \textit{i.e.} first pick up some classes and then select several samples from these classes, we retrain two representative approaches, SPN~\cite{ren2018meta} and M-PL~\cite{li2022platinum}, by using the original and our new setting with different label numbers. Then, we evaluate them by using the semi-supervised inference method with 15 queries and 100 unlabeled samples on \textit{mini}ImageNet and \textit{tiered}ImageNet. As shown in Table~\ref{tab:semiproto}, the accuracy of SPN-N drops sharply, while M-PL-N obtains a slight reduction. {These indicate we cannot correctly infer the performance in the new setting based on the results obtained in the original setting. Therefore, we need a new setting to reevaluate the methods.}

\begin{table*}[tb!]
\centering
\begin{tabular}{cc|cccccc|cccccc}
\hline
      &   & \multicolumn{6}{c|}{\textit{mini}ImageNet}                                            & \multicolumn{6}{c}{\textit{tiered}ImageNet}                                          \\
         \hline
 Methods    &    & \multicolumn{3}{c}{5-way 1-shot} & \multicolumn{3}{c|}{5-way 5-shot} & \multicolumn{3}{c}{5-way 1-shot} & \multicolumn{3}{c}{5-way 5-shot} \\
         \hline
\multicolumn{2}{c|}{Labels Per Class $\longrightarrow$}   & 20        & 50        & 100      & 20        & 50        & 100      & 20        & 100       & 200      & 20        & 100       & 200      \\
\hline
\multirow{3}{*}{SPN} &  Original setting & 29.18     & 42.60     & 48.61    & 40.54     & 53.77     & 60.49    & 50.39     & 54.77     & 54.03    & 61.06     & 64.72     & 64.74    \\
                    & New setting    & 28.03     & 34.24     & 40.44    & 38.37     & 47.94     & 54.78    & 43.60     & 51.02     & 52.22    & 57.68     & 61.62     & 62.50   \\
                    & Drop & 1.15	&\bf 8.36	& 8.17	& 2.17	& 5.83	& 5.71	&\bf 6.79	& 3.75	& 1.81	& 3.38	& 3.10 	& 2.24  \\
\hline

\multirow{3}{*}{M-PL} & Original setting & 40.76 & 44.72 & 45.85 & 54.93 & 58.51 & 61.37 & 42.26 & 43.92 & 44.06 & 57.29 & 57.11 & 58.17 \\
                        & New setting & 39.73 & 43.49 & 45.56 & 53.78 & 57.21 & 59.10 & 42.13 & 43.63 & 43.68 & 56.59 & 56.18 & 57.48 \\
                        & Drop  &1.03 & 1.23 & 0.29 & 1.15 & 1.30 &\bf 2.27 & 0.13 & 0.29 & 0.38 & 0.70 &\bf 0.93 & 0.69 \\
\hline
\end{tabular}
\caption{
\textbf{Accuracy (\%) of $M$-way $K$-shot tasks under the original and new setting.} We show the results of SPN~\cite{ren2018meta} and M-PL~\cite{li2022platinum}   on \textit{mini}ImageNet and \textit{tiered}ImageNet provided by official codes.
The most substantial performance drop for each dataset is highlighted in bold.
}
\label{tab:semiproto}
\end{table*}

\subsection{Benefits of Meta-Learning for SSL}
\begin{figure*}[t]
\centering
  \begin{subfigure}{.495\linewidth}
    \includegraphics[width = .495 \textwidth]{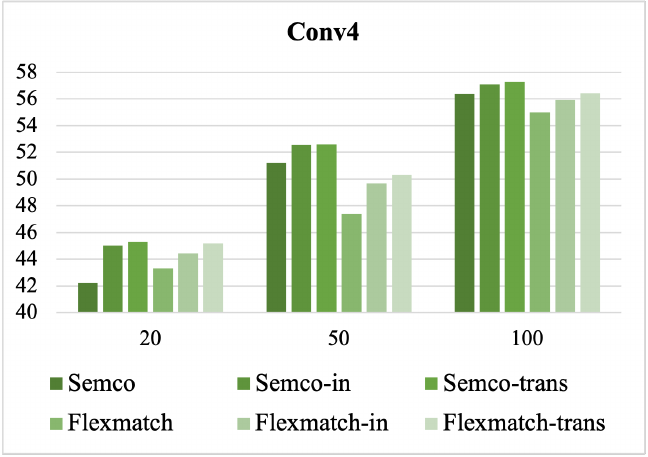}
\includegraphics[width = .495 \textwidth]{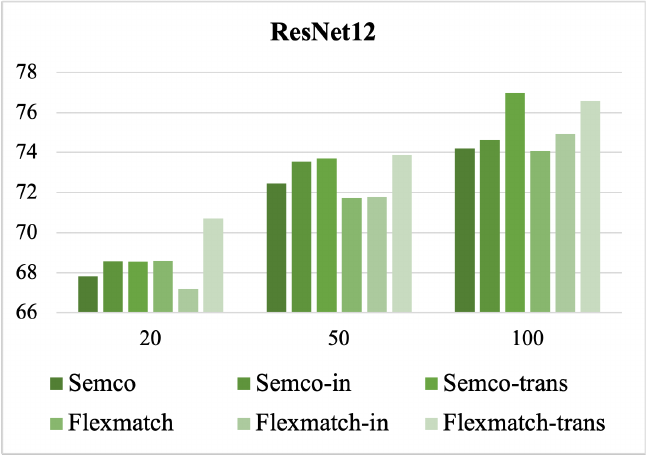}
    \caption{\textit{mini}ImageNet}
    \label{fig:ssl-a}
  \end{subfigure}
  \hfill
\centering
\begin{subfigure}{.495\linewidth}
    \includegraphics[width = .495 \textwidth]{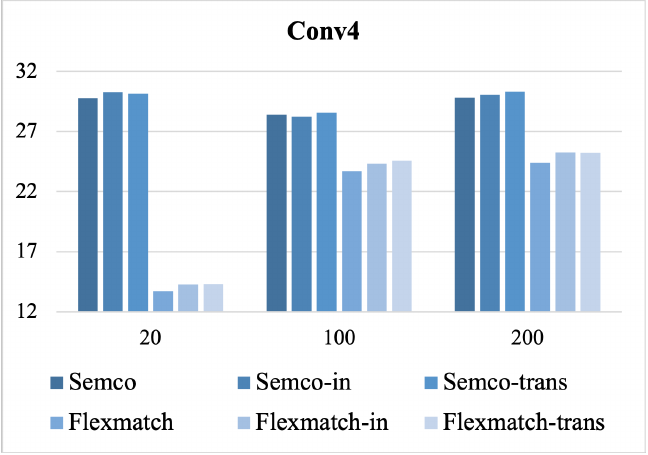}
\includegraphics[width = .495 \textwidth]{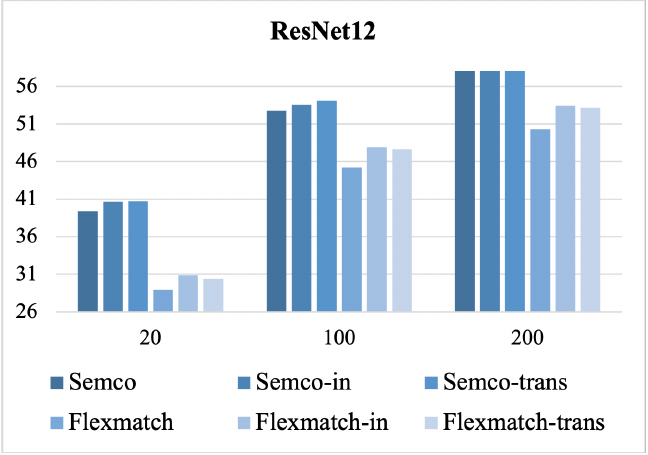}
    \caption{\textit{tiered}ImageNet}
    \label{fig:ssl-b}
  \end{subfigure}
  \hfill
\caption{
\textbf{Accuracy (\%) of SSL methods,} including SemCo, FlexMatch, and our models: SemCo-\textit{in}, SemCo-\textit{trans}, FlexMatch-\textit{in}, and FlexMatch-\textit{trans}. \textit{in} and \textit{trans} represent inductive Proto and transductive EP, respectively. 
}\label{fig:ssl}
\end{figure*}

To investigate the effect of meta-learning on SSL models, we finetune SemCo and FlexMatch by our \ourmodel{}, and evaluate them on the testing set of base classes in \textit{mini}ImageNet, and \textit{tiered}ImageNet. As shown in Fig.~\ref{fig:ssl}, our approaches achieve superior performance than the baselines in most evaluation tasks of three datasets. Specifically, the highest gains are up to 3.04\% in \textit{mini}ImageNet (SemCo-\textit{trans} with 20 labels) and 2.83\% in \textit{tiered}ImageNet (FlexMatch-\textit{trans} with 200 labels) over state-of-the-art SSL algorithms. This indicates the huge potency of meta-learning for SSL, since our approach is an initial and simple attempt to utilize meta-learning for SSL. More techniques can be introduced to combine two topics.

\begin{table}[h]
\centering
\begin{tabular}{l|ccc|c}
\hline
Labels Per Class   & 20      & 50     & 100     & Full  \\
         \hline
PL-Conv4   & 44.46 & 56.22 & 65.84  & 100     \\
FSL-Conv4  & 66.30 & 67.92 & 70.05  & 68.92 \\
\hline
PL-Resnet12  & 69.05 & 76.93 & 81.18  & 100   \\
FSL-Resnet12 & 77.67 & 78.00 & 78.84  & 79.16 \\
\hline
\end{tabular}
\caption{
\textbf{Accuracy (\%) of pseudo-labeling (PL) and 5-way 5-shot  FSL tasks with our SemCo-PLML-\textit{trans}EP on \textit{mini}ImageNet.} 
}
\label{tab:mini-PL}
\end{table}

\subsection{Relation between the Pseudo-labeling Accuracy and Few-Shot Learning Performance}
To investigate the impact of Pseudo-labeling (PL) accuracy, we provide the classification accuracy on the training set of base dataset, and the corresponding few-shot learning (FSL) accuracy in 5-way 5-shot tasks.
As shown in Table \ref{tab:mini-PL}, PL and FSL accuracy scores generally co-improve with the increasing number of labels per class.
Besides, our approach with only 20 labels achieves comparable performance with fully supervised models, even though the PL accuracy is low. 
A better backbone obtains a smaller gap of the full model (C4: 2.62\% \textit{vs.} R12: 1.49\%), indicating an excellent backbone with our approach can improve the accuracy with a few labels.

\begin{table}[h]
    \centering
    \begin{tabular}{ccccccc}
    \hline
                              & Pre   & \multicolumn{2}{c}{LSM} & ND & \multicolumn{2}{c}{\textit{mini}ImageNet} \\
                              &             & TL  & FP  &    & (5,1) & (5,5)  \\
                              \hline
    EP                        &           &           &           &           & 42.31 &  54.79 \\
    \hline
    \multirow{5}{*}{Ours}&\checkmark &           &           &            & 47.56 & 61.39  \\
                              &\checkmark &\checkmark &           &          & 50.48 & 64.51   \\
                              &\checkmark &              &\checkmark &            & 49.85 & 64.57 \\
                              &\checkmark & \checkmark &  \checkmark  &           & 51.86 & 65.59  \\
                              &\checkmark & \checkmark & \checkmark &\checkmark   &\bf 52.59 &\bf 66.30  \\
                              \hline
    \end{tabular}
    \caption{\textbf{Accuracy (\%) on \textit{mini}ImageNet with 20 labels per class.} "Ours" refers to SemCo-PLML-trans-EP, "Pre" refers to Pre-train, "LSM" refers to Learnable Smoothing Module, "TL" refers to Transformer Layer, "FP" refers to Feature Propagation, and "ND" refers to Noise Dropout. Except EP, all transductive models with Conv4 backbone are pre-trained by SemCo. The best values are in bold.}
    \label{tab:ablation}
\end{table}

\subsection{Ablation Study of Our \ourmodel}
To evaluate the key components of our {\ourmodel}, we conduct experiments on \textit{mini}ImageNet with extremely few labels (20) per class. We select the transductive EP with Conv4 as the baseline. {As shown in Table~\ref{tab:ablation}, we first pre-train the model with SemCo and achieve significant improvement (6.69\% and 5.25\% in terms of 5-shot and 1-shot tasks) than EP. Then, we add our Learnable Smoothing Module for feature adjustment and also obtain large gains in both 5-shot (4.20\%) and 1-shot (4.30\%) tasks. Finally, we additionally apply the noise dropout strategy to remove noisy samples during training and observe further improvements in both tasks. Besides, we also test only using a transformer layer or feature propagation for feature adjustment.} However, both of them achieve worse performance than our LSM, clearly demonstrating our model's effectiveness.

\begin{table}[htb!]
\centering
\begin{tabular}{cc|ccc}
\hline
Training & Backbone & \multicolumn{3}{c}{5-way 1-shot} \\
\hline
\multicolumn{2}{c|}{Labels Per Class $\longrightarrow$} & 20   & 50   & 100     \\
\hline

 EP                &Conv4    & 42.31  & 46.78  & 49.01\\
PL-EP    & Conv4      &49.85(+7.54)  &52.40(+5.62) &54.82(+5.81) \\
\ourmodel-EP  & Conv4  &\bf52.59(+10.28) &\bf54.03(+7.25)  &\bf 56.23(+7.22)    \\
\hline
 EP                &ResNet12    & 44.77  & 50.63  & 54.93 \\
PL-EP  & ResNet12   &61.45(+16.68)  & 61.98(+11.33) & 62.01(+7.08) \\
\ourmodel-EP  & ResNet12 &\bf 64.45(+19.68)  &\bf 64.78(+14.15) &\bf64.40(+9.47) \\
\hline
\end{tabular}
\caption{\tr{\textbf{Accuracy (\%) of $M$-way $K$-shot tasks on \textit{mini}ImageNet.} PL-EP and \ourmodel-EP use SemCo in the pre-training stage, and the inference method is \textit{trans}-EP. 
The best values are in bold.}
}
\label{tab:vanilla-pl}
\end{table}

\subsection{\tr{Benefits of Pseudo-Labeling}}
\tr{To demonstrate the benefits of our pre-training method based on pseudo-labeling, we conducted experiments on \textit{mini}ImageNet. As shown in tabel~\ref{tab:vanilla-pl}, we replaced the pre-training method in EP with our pseudo-labeling training stage (PL-EP) and achieved significant improvement in all tasks compared to the original EP. This indicates our pseudo-labeling can easily incorporated with other SSFSL methods to improve the performance with fewer labels. However, the simple incorporation still suffers from the problem of noisy samples. Thus, we proposed finetuning with noise suppression to alleviate this issue and our \ourmodel{}-EP achieves significant improvements under all conditions. Specifically, the highest gains are up to 10.28\% with Conv4 as the backbone (20 labels per class) and 19.68\% with ResNet12 as the backbone (20 labels per class).}

\begin{table}[!htb]
\centering
\begin{tabular}{c|ccccc||c}
\hline
 $\alpha_d$ & 0     & 0.1  & 0.2   & 0.3   & 0.5 & LSM  \\
\hline
(5,5)        & 65.74 &\bf 66.3 & 66.14 & 66.18 & 65.78 & 65.59 \\
\hline
\end{tabular}
\caption{\textbf{Accuracy (\%) of 5-way 5-shot tasks  on \textit{mini}ImageNet~\cite{ravi2016optimization} with 20 labels per class}. We show the results with different trade-off parameter $\alpha_d$, based on the SemCo~\cite{nassar2021all} pre-training and \ourmodel-EP finetuning methods. The backbone is Conv4. The best values are in bold. 
}
\label{tab:drop}
\end{table}

\begin{table}[!htb]
\centering
\begin{tabular}{c|cccccc||c}
\hline
 $\gamma$          & 0     & 0.1   & 0.2   & 0.5   & 1     & 2     & Pre\\
\hline
(5,5) & 65.69 &\bf 66.3  & 65.06 & 65.62 & 64.77 & 64.39 & 61.39  \\
\hline
Base & 44.28 &\bf 45.28 & 45.23 & 44.61 & 44.61 & 43.33 & 42.23\\
\hline
\end{tabular}
\caption{\textbf{Accuracy (\%) of 5-way 5-shot tasks and base classification on \textit{mini}ImageNet~\cite{ravi2016optimization} with 20 labels per class}. We show the results with different trade-off parameter $\gamma$, based on the SemCo~\cite{nassar2021all} pre-training and \ourmodel-EP finetuning methods. "Pre" refers to Pre-training. The backbone is Conv4. The best values are in bold. 
}
\label{tab:gamma}
\end{table}

\subsection{Ablation Study for Hyperparameter}
One of our main hyperparameters is the dropout rate $\alpha_d$ (in the noise dropout strategy of \S\ref{sec:fns}). We observe the performance variation on the 5-way 5-shot task on \textit{mini}ImageNet with 20 labels, by using different $\alpha_d$. As shown in \cref{tab:drop}, when we only resort to our support set and do not dropout any samples ($\alpha_d=0$), we achieve a slight improvement compared with the original set (LSM). Then, we increase $\alpha_d$ from 0.1 to 0.3 and obtain similar improvements. This indicates that dropping out a small number of samples will improve performance. When we increase it to 0.5, the accuracy reduces. This means removing too many samples will hurt the final training performance, because of the lacking of various samples. Therefore, we experimentally set our dropout rate $\alpha_d$ with a small value of 0.1.

The other main hyperparameter is the trade-off parameter $\gamma$ (in the training loss of \S\ref{sec:fns}) of the base and few-shot learning (FSL) classification losses. To investigate its impact for both two tasks, we use different $\gamma$ for finetuning. As shown in \cref{tab:gamma}, when we do not use base classification loss ($\gamma$=0), we improve both FSL performance compared with the pre-training model (65.69\% vs. 61.39\%) and the base classification accuracy (44.28\% vs. 42.23). Then we try different $\gamma$ and find when $\gamma=0.1$, we achieve the best performance in terms of both two tasks. Thus, $\gamma$ is experimentally set as 0.1.

\subsection{Limits and Discussion}
{There are still some shortcomings to address. Firstly, achieving performance equivalent to models trained with full labeled data still necessitates a relatively large number of labels, typically around 100 labels, leaving substantial room for improvement in tasks with fewer labels, such as those with 20 or 50  (see Table \ref{tab:semi-fsl}). Secondly, experimental results (\textit{e.g.} Table \ref{tab:mini}) suggest that the efficacy of the final few-shot learning method is closely intertwined with the effectiveness of semi-supervised learning. From this standpoint, enhancing performance further entails designing better training mechanisms to fully exploit meta-learning for the training of semi-supervised models.}


\section{Conclusion}
We proposed a novel semi-supervised setting for FSL to explore effective training solutions with a few labels. Under this setting, we found that most FSL approaches suffered from huge performance degradation with the reduction of labels. To alleviate this issue, we presented a simple yet effective two-stage learning framework, called meta-learning with pseudo-labeling (\ourmodel), which successfully utilizes the advantages of SSL and FSL for effective learning from a few labels. We integrated \ourmodel{} into three SSL and FSL models. Extensive results on two datasets have demonstrated that our approach not only provided significant improvement over FSL baselines but also achieved superior performance than SSL models. In a sense, \ourmodel{} is an interface between FSL and SSL, and we believe it has a large potential to facilitate both two topics.

\appendix
\section{Symbols and Abbreviations}
{As shown in Table~\ref{tab:symbol}, we have compiled the main symbols and abbreviations used throughout the text for easier reference.}
\begin{table}[h]
    \centering
    \begin{tabular}{c|l}
        \hline
        \bf Symbol/ &  \multirow{2}{*}{\bf Definition}  \\
        \bf Abbreviation &   \\
        \hline
        FSL        & few-shot learning  \\
        IFSL        &inductive few-shot learning          \\
        TFSL        & transductive few-shot learning        \\
        SSFSL        & Semi-supervised few-shot learning         \\
        *-PLML        & our method           \\
        *-N        & our new setting           \\
        \hline
        $\mathcal{D}_\mathit{base}$        & base set          \\
        $\mathcal{D}_\mathit{novel}$        & novel set          \\
        $\mathcal{D}_\mathit{l}$        & labeled set in the semi-supervised setting   \\
        $\mathcal{D}_\mathit{u}$        & unlabeled set in the semi-supervised setting   \\
        $\mathcal{D}_\mathit{t}$        & testing set in the semi-supervised setting   \\
        $\mathcal{D}_\mathit{p}$        & pseudo-labeling dataset   \\
        $f_\mathit{ssl}$        & Semi-supervised learning model   \\
        $f_\mathit{fsl}$        & few-shot learning model   \\
        $\phi$        & feature extractor   \\
        $c_f$        & M-way classifier   \\
        $c_b$        & base classifier   \\
        $\mathcal{Y}_s$        & predicting label of $c_f$            \\
        $\mathcal{Y}_b$        & predicting label of $c_b$            \\
        \hline
        $\mathcal{T}$        & transformer operation           \\
        $Z$ & Original feature matrix \\
        $Z_t$ & feature matrix adjusted by $\mathcal{T}$  \\
        $A_{ij}$ & adjacency matrix element  \\
        $d_{ij}^2$ & squared Euclidean distance  \\
        $\sigma^2$ & variance  \\
        $P$ & propagator matrix \\
        $I$ & identity matrix \\
        $L$ & laplacian matrix \\
        $D_{ii}$ & the degree of node $i$ \\
        $\alpha_p$ & scaling factor \\
        $Z_p$ & feature matrix adjusted by propagation\\
        $z_{ai}$ & feature of the anchor sample \\
        $\alpha_d$ & dropout rate \\
        \hline
    \end{tabular}
    \caption{List of Symbols and Abbreviations}
    \label{tab:symbol}
\end{table}

\bibliographystyle{ieee_fullname}
\bibliography{egbib}

\vfill

\end{document}